# Knife and Threat Detectors

**David A. Noever**
PeopleTec, Inc.
Huntsville, AL
david.noever@peopletec.com

**Sam E. Miller Noever**
Independent Consulting
Huntsville, AL
sam.igorugor@gmail.com

**ABSTRACT**

Despite rapid advances in image-based machine learning, the threat identification of a knife wielding attacker has not garnered substantial academic attention. This relative research gap appears less understandable given the high knife assault rate (>100,000 annually) and the increasing availability of public video surveillance to analyze and forensically document. We present three complementary methods for scoring automated threat identification using multiple knife image datasets, each with the goal of narrowing down possible assault intentions while minimizing mis-identifying false positives and risky false negatives. To alert an observer to the knife wielding threat, we test and deploy classification built around MobileNet in a sparse and pruned neural network with a small memory requirement (< 2.2 megabytes) and 95% test accuracy. We secondly train a detection algorithm (MaskRCNN) to segment the hand from the knife in a single image and assign probable certainty to their relative location. This segmentation accomplishes both localization with bounding boxes but also relative positions to infer overhand threats. A final model built on the PoseNet architecture assigns anatomical waypoints or skeletal features to narrow the threat characteristics and reduce misunderstood intentions. We further identify and supplement existing data gaps that might blind a deployed knife threat detector such as collecting innocuous hand and fist images as important negative training sets. When automated on commodity hardware and software solutions one original research contribution is this systematic survey of timely and readily available image-based alerts to task and prioritize crime prevention countermeasures prior to a tragic outcome.

**INTRODUCTION**

As machine learning (ML) gets better at mimicking the capabilities of human experts, more crime prevention and forensic investigations will transform from specialty disciplines into more routine tasks for generalists. As outlined in Figure 1, this paper explores the strengths and weaknesses of current ML for active threat detection. In addition to traditional forensics based on imagery (fingerprints, ballistics, etc.), audio (voice recognition), and text (handwriting, style recognition), we highlight new investigative tools that only ML can execute. The ability for object detectors and scene classifiers to automate threat identification in video and still frames plays an important role in many preventive crime settings, such as schools, churches, concerts and public gatherings. In 2018, there were 123,253 aggravated assaults in the US where knives or other cutting instruments were used (Statista, 2019). According to the Bureau of Justice Statistics (NCVRW, 2018), 12 out of every 1000 people were victims of simple assault. Law enforcement and nursing present increased risks by profession. In 2015, 33% of aggravated assault victimizations were committed with a knife, 23% with a firearm, and 11% without a weapon. For many of these public assaults outside the home where surveillance and closed-circuit TV (CCTV) may already exist, the automated notification and alerts needed for active and timely threat identification has proven too cumbersome previously to analyze and archive without machine assistance. Nearly 54% of male victims and 37% of female victims were attacked at either a commercial place, parking lot, public area or school (NCVRW, 2018).

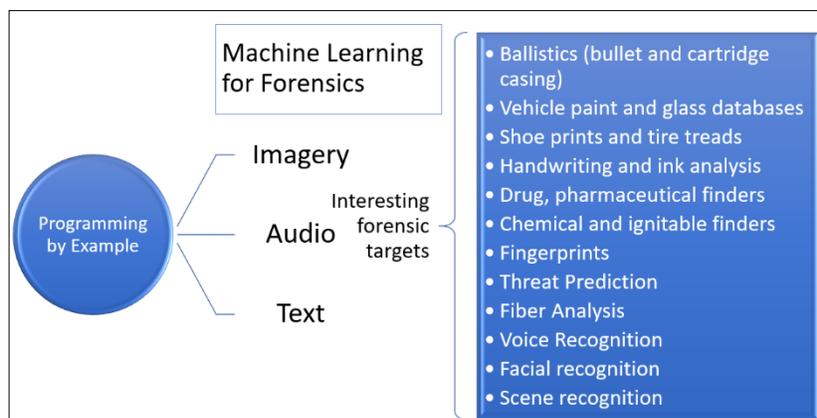

*Figure 1. Traditional forensic tasks that machine learning can present new tools*

As noted by Buckchash, et al. (2017), detecting knives offers a machine learning challenge for identifying both the object and threat intent. As the highlighted object (Figure 2), a generic knife blade itself presents many variations in rotation angle, size, shape and blade or handle textures, particularly for different intended uses such as paring or chopping. For the present purposes we limit the definition to a traditional large knife that might qualify in many stabbing incidents with greater than an 8-inch blade length. Unlike handguns, rifles and other threatening objects, the lack of a uniform model catalog for knives translates into a much larger set of potential search choices. The threat identification itself presents a further challenge when false positives and negatives may overwhelm the real cases of interest and where inferred intent may prove critical. For instance, any knife in an elementary school (outside of the cafeteria) should appear as threatening instance. A knife however shown in a craft workshop or kitchen may present threatening intents only if held in specific ways, like the overhand or downward stab characteristic of an attack in progress.

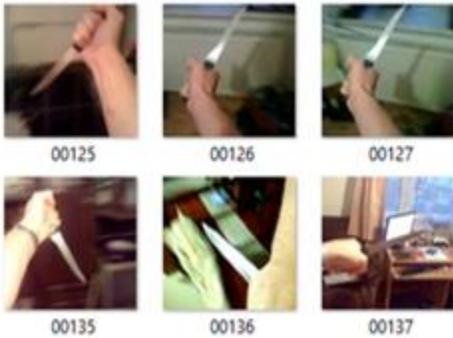

Figure 2 Knife dataset with hand (adapted from Shekhar, 2020)

**Previous Work.** Prior to 2012 previous work to build and test robust knife detectors had relied on classical computer vision techniques before deep learning began to dominate image analysis competitions and achieve expert-level results. Computer vision methods included edge detection, image skeletonization, thresholding and other techniques for pre-processing features that could link shape, size or pose to a conclusion about threat intents. Zywick, et al (2011) for instance applied Haar filters (as popularized for facial feature detection) for their multi-scale detection. In their conclusions, the authors noted that the algorithm executed slowly (3 s) in iterations that detected knives in only 45% of the true cases and 84% of the negative cases without actual knives in the image. Their accomplishment in assembling a large, labeled dataset (>10,000 cases) however motivated an updated revisit using deep learning methods (Figure 3). Later work has specialized some detection methods to specific parts of the spectrum (X-ray, visible, black and white), camera (video, noisy CCTV) and motion tracking. It remains a notable challenge to marshal the correct datasets for size and diversity with the most current end-to-end detection systems that can infer both likely classifications, detection localizations, segmentation of multiply connected objects (hand and knife), and finally position or pose to infer contextual information like threatening intent.

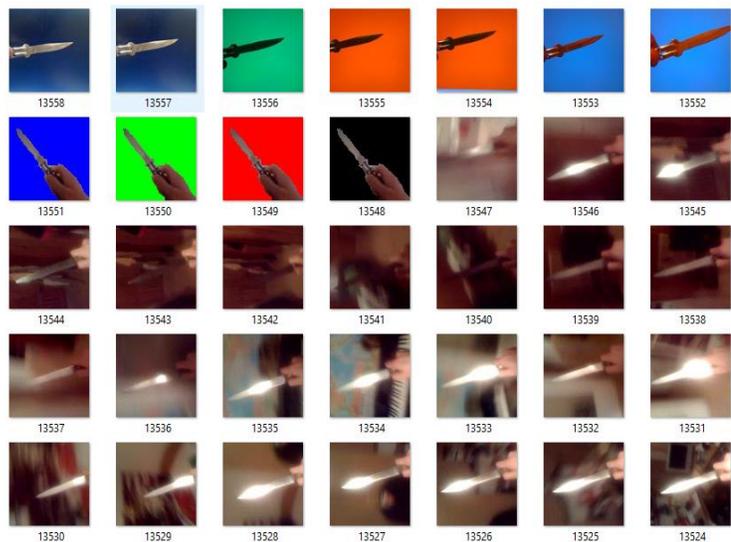

Figure 3. Video frame capture dataset adapted from Matiolanski, et al. (2016) and Grega, et al (2016)

**Original Contributions.** To the authors' knowledge, this work is the first to address the combined classification, detection and pose challenges of knife and threat intent using a modern and comprehensive deep learning framework. We demonstrate high accuracy (>95%) and diverse algorithms on multiple knife datasets and supplement multi-classifications with our own independent datasets that exclude likely false positives such as hands, fists and palms or disembodied knives that otherwise should appear as non-threatening with appropriate scene context. This work also pioneers a video-capable frame rate for rapid detection even in low-powered devices like a browser that otherwise needs no connectivity to larger computing power. This edge case suggests new opportunities to deploy machine learning in key crime detection and prevention scenarios like schools and community gatherings that might lack connectivity or resources expected in a large urban network operations center (NOC). Crisis management for these instances has traditionally depended on physical security devices and staffed inspection points to identify threats, often at much greater expense in manpower, money and time, while also inconveniencing the vast majority of innocent or non-threatening by-standers. Throughout the analysis, our approach will highlight the unique aspects of knife detection for its widely varying feature space (size, shape, lighting, texture). This work handles many of the more challenging algorithmic problems by adopting a popular choice to ignore expertly designed features or expected image circumstances in favor of simply learning by example with a larger and more diverse dataset that reflects a self-evolving hierarchy of features built from samples and experience. As a practical matter, this approach has offered the research community a consistent and pixel-based approach to achieve expert-level classification (>95%) and detection localizations (>75%) across a wide scope of image tasks.

**METHODS**

We solve the knife threat identification as both a classification and detection problem. The classification version highlights whether a given video frame or image contains a knife and hand, or just a hand without knife or neither. The detection version localizes the knife and hand in the same image and assign both independent bounding boxes and probable detection certainties from a pixel level. As a supplementary method for detecting intent, we also train a pose model that infers key skeletal points to distinguish the knife position with a grasping hand from a simple empty fist. One hypothesis to test is how well this combination can aid decision-making for threats. For instance, a cooking show may feature many grasped and countertop knives but not support a conclusion for threat. Similarly, a human observer of video might conclude differently from the overhand knife as being more threatening a posture than the basic slicing motion seen in a benign kitchen scene.

**Datasets.** Matiolanski, et al. (2016) and Grega, et al (2016) have published a dataset of 12,799 100x100 color images with 27.8% labeled as positive knife cases taken from video frames. They have specialized their object detector for knives to the closed-circuit TV case which can address the resolution and enhancement challenges from video. We train the multi-class problem of "A: threat, B: no threat without hand, or C: no threat with hand" using transfer learning and MobileNet architecture. The need to introduce two negative cases (B and C) arises because of many false positives for threats if detecting hands only, since typically two objects are featured in the knife datasets: the weapon and the grasping hand. This step disaggregates the class assignments. We trained a classifier to give the probability of a given video frame to include a threat (knife) or not (no hand or no knife). We train the classifier for 50 complete passes (epochs) through the 3559 knives, an equal number of no threat instance. The classifier is built with a batch size of 16, and gradual learning rate (0.001) since we want to keep the pre-trained network architecture from overly disruptive iterative cycles. We supplemented the hand images from Google images and included a mix of negative poses for palms, fists and position angles (147 total).

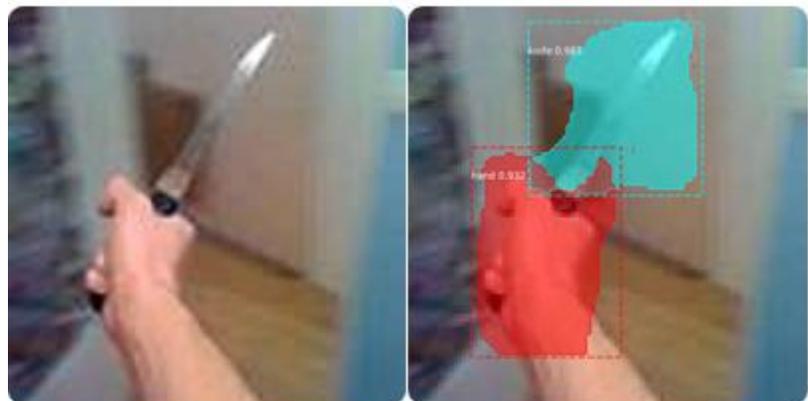

*Figure 4 Before and after images from MaskRCNN. The image algorithm learns to segment the hand (red) and knife (blue) separately in the same image with assigned confidence levels for pixel masks.*

Recent authors have published additional knife datasets, either absent context (Singh, 2020), or hand-held in close-up for active threat detection (Shekhar, 2020). Since both cases offer less than a thousand examples, these datasets can also benefit from the previously cited transfer learning strategies but also with data augmentation. Compared to the CCTV examples, these higher resolution images were labelled with bounding boxes and segments for both the hand and knife. This more detailed labeling enables masking algorithms and detection algorithms (MaskRCNN) to locate the position and certainty for the hand and knife separately in the same image.

**Algorithms.** We test the applied success of multiple neural network designs including MobileNet for classification, MaskRCNN for detection and PoseNet for positioning skeletal points. In all these cases, we depend on transfer learning from a pre-trained starting point which can take advantage for existing features such as shape, texture and edge detection learned from tens of millions of unrelated observations to our use case. A notable outcome of this approach can assist augmenting smaller datasets (<1000 samples) and networks capable of working on edge devices like disconnected CCTV hardware without a network hub or large computing resource. One limitation of this broad approach to scoring knife identification is the final decision-making layer, where a hierarchical classifier might issue alerts or notifications that an imminent threat was flagged. Most of these decision trees or ensembled voting algorithms depend on expertly labeled datasets which humans have collectively assigned an actionable outcome to a certain set of inputs. While not technically challenging, this final actionable notification layer depends on data presently unavailable to the open research community. We train MaskRCNN on multiple NVIDIA V100 GPUs over 200 epochs after individually labeling hand and knife localizations for 500 instances as published without labels by Shekhar (2020). Our labeling benefited from the "labelImg" open source graphical image annotation software on Github (Tzutalin, 2015) and tagged annotations were saved as Pascal VOC format. For all datasets, our approach follows the traditional training and testing split commonly seen for image-based problems with 70% training, 15% validation and 15% testing divisions. We did not systematically perform cross-fold validation with random subsets of images but observe from previous experience that transfer learning can compensate for overfitting smaller image datasets and typically avoid the pitfalls of memorizing few examples with many free parameters. In other words, the millions of images used to build initial feature detectors in pre-training can prime the detector and avoid suboptimal overfitting. Multiple researchers have documented the outperformance of this transfer learning strategy compared to initiating network weights randomly and trying to train from scratch.

## RESULTS

We summarize the accuracy (and entropy loss function) as a function of training epochs in Figures 6-7, Supplementary Material. For classification using MobileNet, the large knife datasets published by Matiolanski, et al. (2016) and Grega, et al (2016) yield greater than 95% accuracy for correct identification of knife, hand or neither among a three-class test set. For detection using MaskRCNN, the smaller dataset published by Shekhar (2020) provide a similar high threshold for correct identification of both hand and knife in the same images with greater than 90% certainty. We scored an overall loss function of 0.1 among the 20% test images withheld from training epochs (Figure 4, Figure 8, Supplementary Material). A significant feature of this segmentation or masking for the hand and knife together can define future decision tree rules that assist the threat determinations. For example, a simple set of positioning rules can prioritize an overhand strike if the bounding box for the hand appears above the knife compared to a less threatening pose. We further trained the PoseNet algorithm (Figure 5) for putting key anatomical points on the arm and hand holding the knife but found the single wrist point as the major identifier and provide less information than one might expect from a full skeletal view in surveillance or CCTV footage.

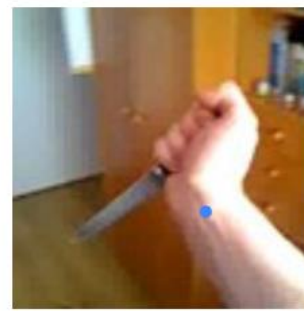
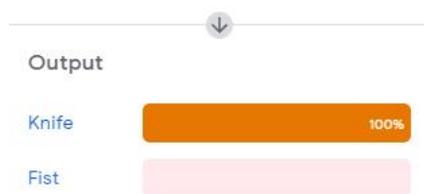

*Figure 5. PoseNet wrist identification (blue skeletal point) and scene classification from video for distinguishing knife from fist in threat identification*

**CONCLUSIONS AND FUTURE WORK**

While our work introduces knife pose to infer threats, the treatment can benefit from additional study such as multiple viewpoints and sensor fusion for deeper inspection of context. A shortcoming of threat identification from image data alone is the high risk attached to false negatives and positives where complacent alerts may miss a true incident or flag an innocent interaction as threatening. Typically, this poor tuning gets systematic treatment using larger and more diverse datasets but can stubbornly resist attempts to minimize real-world incidents given the large feature space of exceptional cases. Compared to detecting other threats like firearms, every kitchen or workshop has knives that when rendered from even short videos can yield thousands of training examples without tedious collection steps. An important strategy for addressing these acknowledged shortcomings can rely on unsupervised anomaly detection, one promising approach that designs a built-in robustness to identify any behavior or object out of the ordinary if using a strong baseline for normal vs. abnormal cases. A final supplementary candidate for improved detection would fuse the visible detections with other parts of the available crime prevention chain, such as X-ray, metal detection, and thermal or night vision where a high risk may justify its use for concerts, airports, rallies or other stadium-scale gatherings. While schools and churches may balk at the undertaking, the availability of low-cost hardware when supplemented by modern machine learning may offer alternative methods that outperform manual labor and expert inspection strategies with automated alerts during on- and off-hours. We anticipate that other threat detectors for objects such as handguns, rifles, illicit cargo, and wired vests may benefit from a similar image-based, deep learning algorithm as the research community collates larger real-world or synthetic training datasets.

**ACKNOWLEDGEMENTS**

Żywicki, M., Matiolański, A., Orzechowski, T. M., & Dziech, A. (2011). Knife detection as a subset of object detection approach based on Haar cascades. In Proceedings of 11th International Conference "Pattern recognition and information processing (pp. 139-142).

**SUPPLEMENTARY MATERIAL**

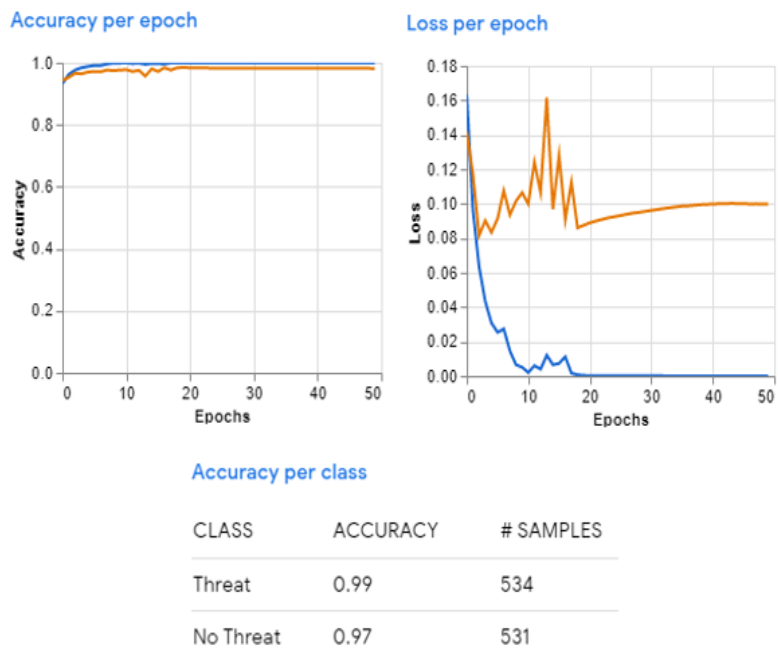

*Figure 6 Training accuracy and loss per epoch. Overall threat vs. no threat identification based on MobileNet and approximately balanced datasets for 1000 total cases. Blue lines show training passes and orange lines show validation values.*

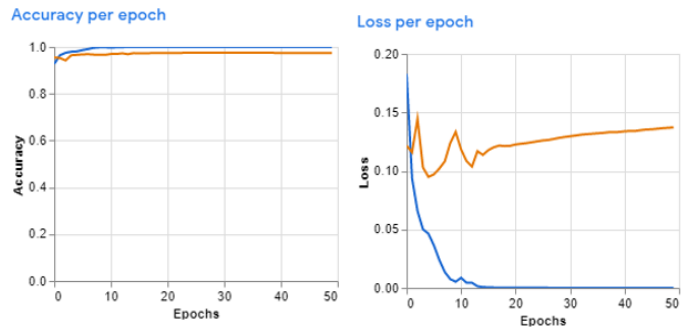

*Figure 7 Training accuracy and loss per epoch. Overall threat vs. no threat with hand data supplemental identification based on MobileNet and approximately balanced datasets for 1000 total cases. Blue lines show training passes and orange lines show validation*

| Loss Score | Training | Validation |
|---|---|---|
| Overall | 0.1849 | 0.9966 |
| RPN Class | 0.0009 | 0.0071 |
| RPN Bounding Box | 0.0547 | 0.3229 |
| MRCCN Class | 0.0195 | 0.0779 |
| MRCNN Bounding Box | 0.0238 | 0.1355 |
| MRCNN Mask | 0.0860 | 0.4533 |

*Figure 8 Training and validation scores for MaskRCNN models with class bounding box, and masking.*